\documentclass[runningheads]{llncs}
\usepackage[T1]{fontenc}
\usepackage{amsmath}
\usepackage{booktabs} 
\usepackage{multirow}
\usepackage{color}
\usepackage{caption}
\usepackage{graphicx}

\begin{document}
\title{LE2Fusion: A novel local edge enhancement module for infrared and visible image fusion}
\titlerunning{LE2Fusion}
%
\author{Yongbiao Xiao \and
Hui Li\thanks{Corresponding author} \and
Chunyang Cheng \and
Xiaoning Song}

\authorrunning{Y. Xiao et al.}
\institute{Jiangsu Provincial Engineering Laboratory of Pattern Recognition and Computational Intelligence, School of Artificial Intelligence and Computer Science, Jiangnan University, 214122, Wuxi, China\\
\email{lihui.cv@jiangnan.edu.cn}\\
}
\maketitle              
\begin{abstract}
Infrared and visible image fusion task aims to generate a fused image which contains salient features and rich texture details from multi-source images. However, under complex illumination conditions, few algorithms pay attention to the edge information of local regions which is crucial for downstream tasks. To this end, we propose a fusion network based on the local edge enhancement, named LE2Fusion. Specifically, a local edge enhancement (LE2) module is proposed to improve the edge information under complex illumination conditions and preserve the essential features of image. For feature extraction, a multi-scale residual attention (MRA) module is applied to extract rich features. Then, with LE2, a set of enhancement weights are generated which are utilized in feature fusion strategy and used to guide the image reconstruction. To better preserve the local detail information and structure information, the pixel intensity loss function based on the local region is also presented. The experiments demonstrate that the proposed method exhibits better fusion performance than the state-of-the-art fusion methods on public datasets.
\end{abstract}
\section{Introduction}

Image fusion is a technique to fuse images acquired by sensors of different types into a single informative image.
As an important image enhancement technique, it has a wide range of applications in remote sensing~\cite{li2018fusing}, object classification~\cite{gao2018object}, \textit{etc}. In this work, we focus on the infrared and visible image fusion.

Due to the complementary characteristics of the infrared and visible image fusion and the development of deep learning, many learning-based fusion methods have been proposed, and the performance of these methods has been greatly improved. Some methods use a single network structure to directly generate the fused image. Although promising results can be obtained, the simple network structure may not effectively preserve useful details. Therefore, in order to retain more details and salient features in the fused images, some methods separately design enhancement modules and include them in the fusion network.\par
\begin{figure}[t]
\vspace{-0.2cm}
\centering
\includegraphics[width=\textwidth]{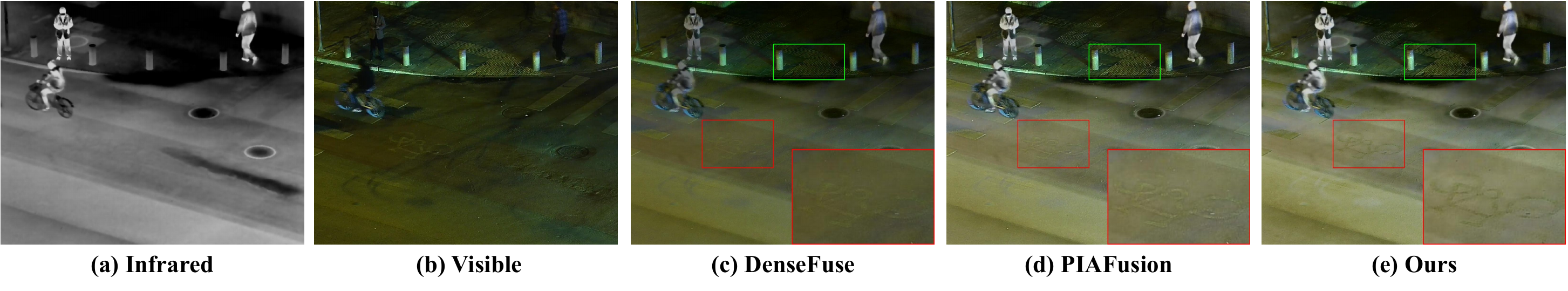}
\vspace{-0.2cm}
\caption{An example of infrared and visible image fusion. Our method preserves edge information and details of bicycle mark.} \label{fig1}
\vspace{-0.5cm}
\end{figure}

On the other hand, few learning-based algorithms pay much attention to the local edge information and salient features under complex illumination conditions.
As shown in Fig.~\ref{fig1}, compared with the illumination-aware method PIAFusion~\cite{tang2022piafusion}, our result can fully present the sign of the bicycle on the ground.
This issue indicates that the enhancement module in the PIAFusion are not robust.

To address the aforementioned problem, we propose a novel fusion network for the infrared and visible image fusion based on the local edge enhancement module, which is termed as LE2Fusion.
In LE2Fusion, a local edge enhancement module is designed to simultaneously enhance the local edge information and preserve the salient features.
After that, a multi-scale residual attention module is applied into feature extractor and the rich features of source images can be obtained.
Besides, a set of enhancement weights are utilized to guide the fusion stage and the reconstruction of the fused image.
Finally, we innovatively design a pixel intensity loss function based on the local region, which enables the network to preserve more local structural features and thermal radiation information from the infrared modality. In this way, regardless of the illumination conditions, our method can produce promising fusion results. In summary, the major contributions of this study are summarized as follows:\par
\begin{itemize}
\item A local edge enhancement module is devised to extract the local edge intensity information. A set of enhancement weights generated by this module are used to guide the fusion and reconstruction process of the final output.

\item A multi-scale residual module based on the attention mechanism is introduced in the feature extractor to obtain more rich deep features of source images.

\item We design a new pixel intensity loss function based on the ideas of preserving the structural features and pixel intensity in the local region.

\item Qualitative and quantitative experiments on multiple infrared and visible image fusion benchmarks demonstrate the superiority of the proposed method.

\end{itemize}
\par

\section{Related work}
In this section, we first review the image fusion algorithms without the enhancement module.
After that, we briefly introduce some methods based on the enhancement module.

\textbf{Algorithms without the enhancement module:} Without applying the enhancement modules, current algorithms only improve and optimize the network itself. DenseFuse~\cite{li2018densefuse} uses dense blocks during the encoding phase to extract image features. In addition to the dense connections, Li \textit{et al.} introduce a multi-scale encoder-decoder network and the nested connections to extract more comprehensive features of source images (RFN-Nest~\cite{li2021rfn}). 
On the other hand, IFCNN~\cite{zhang2020ifcnn} is a representative general fusion network. 
In this work, the convolutional layers are used to extract salient features of source images. 
Then, the combined convolutional features are reconstructed to yield the fused images.
Although these methods can achieve promising results, the design of the network structure is too simple to fully extract the robust features of the images. 
Besides, the coarse-grained loss function design cannot meet the requirements to focus on the the details and structural information of source images.

\textbf{Algorithms with the enhancement module:} Since the use of a single network cannot achieve the desired result, some approaches introduce additional modules to enhance the fusion results. SeAFusion~\cite{tang2022image} designs a gradient residual dense block to enhance fine-grained spatial details. DIVFusion~\cite{tang2023divfusion} designs two modules to remove the illumination degradation and enhance the contrast and texture details of the fused features respectively. In MUFusion~\cite{cheng2023mufusion}, a new memory unit architecture based on the intermediate fusion results collected in  the training phase is introduced to further supervise the fusion process.
However, although some modules are used in their methods, the enhancement performance is not promising.
Thus, we need fine-grained design to improve the current enhancement-based methods.

\section{Method}
In this section, the proposed fusion network is presented. Firstly, we give the details of our framework~\cite{fu2023mdranet}. Then, the loss functions for training phase are introduced. Finally, the detail settings of our network will be given.
\begin{figure}[!htbp]
\vspace{-0.5cm}
\centering
\includegraphics[width=0.8\textwidth]{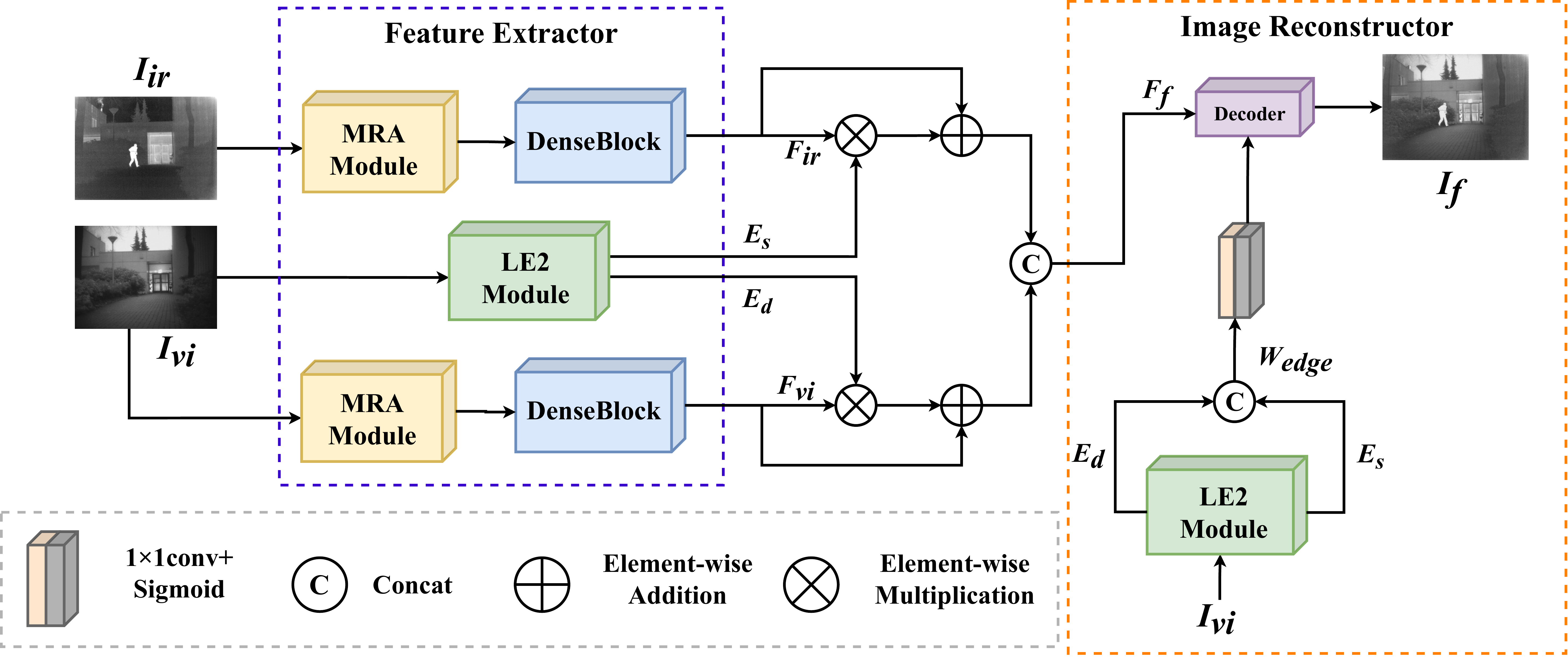}
\vspace{-0.2cm}
\caption{The framework of the proposed method.} \label{fig2}
\vspace{-1.3cm}
\end{figure}

\subsection{Framework}
As we discussed, most image fusion methods ignore the local region edge information in the feature extraction process. To solve this problem, we design a fusion network based on the local edge enhancement module. The framework is shown in Fig.~\ref{fig2}. Our method combined with two architectures ("Feature Extractor" and "Image Reconstructor") in which the LE2 is utilized in these two module simultaneously. In feature fusion strategy, the attention and the concatenation are introduced to preserve the complementary features.

\subsubsection{Local edge enhancement module}
\begin{figure}[!htbp]
\centering
\vspace{-0.4cm}
\includegraphics[width=0.8\textwidth]{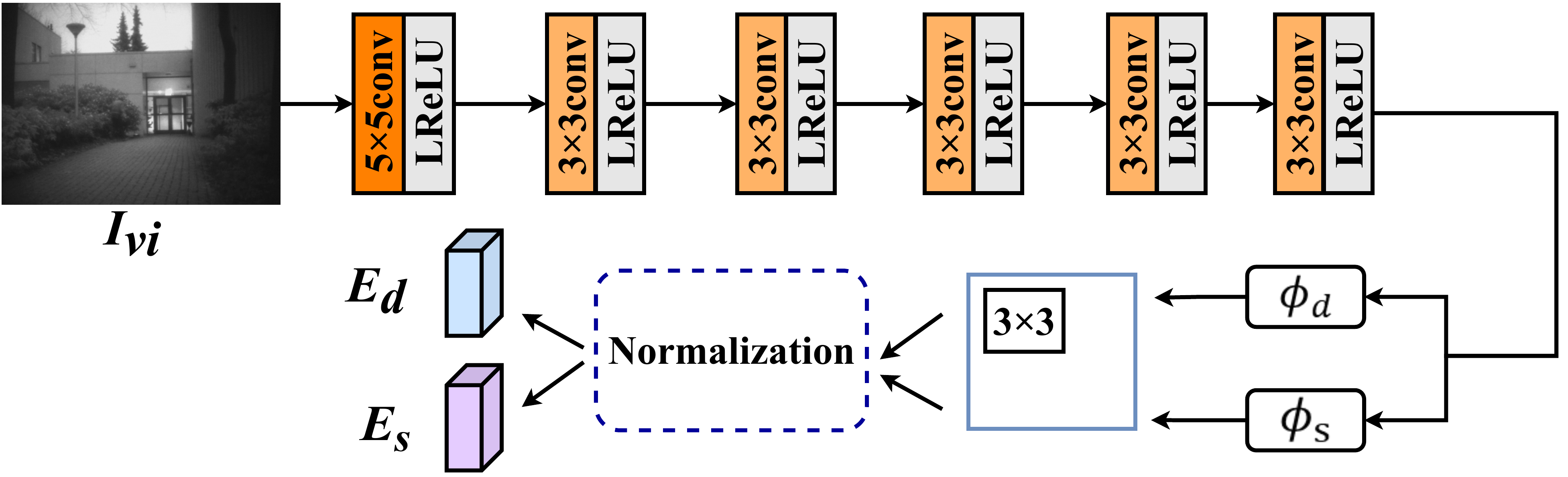}
\vspace{-0.4cm}
\caption{The detailed structure of the local edge enhancement module (LE2).} \label{fig3}
\vspace{-0.5cm}
\end{figure}

As shown in Fig.~\ref{fig3}, considering the imbalance of illumination will degrade the image feature extraction, a local edge enhancement module is proposed to avoid this drawback. The LE2 module contains five convolutional layers and one normalization operation to generate the illumination-aware matrices ($E_d$ and $E_s$).

Given a visible image $I_{vi}$, the LE2 module is formulated as:
\begin{equation}
\left\{\phi_d, \phi_s \right\}=N_{edge}(I_{vi})
\end{equation}
where $N_{edge}$ denotes the edge enhancement network, $\phi_d$ and $\phi_s$ are the extracted edge intensities and salient features.

Besides, compared with the general methods that only focus on the pixel intensity of the source images, the features of the $3\times3$ region extracted by our method not only focus on the image pixel intensity, but also cover the local structure of the images, making the overall image features more robust.
This process is formulated as:
\begin{equation}
L_c=N_{local}(\phi_c), \quad c\in \left\{d, s \right\}
\end{equation}
where $N_{local}$ represents the network used to extract the features in the local region, $L_d$ and $L_s$ are the edge intensity and the salient features in the local region, respectively.

Meanwhile, we use a simple normalization function to calculate the final weights which also indicate the local region edge information from the source images, the formula is defined as follows,
\begin{equation}
E_c=\frac{L_c}{\sum_{i\in \left\{d, s \right\}} L_i}, \quad c\in \left\{d, s \right\}
\end{equation}
where $E_d$ and $E_s$ indicate the edge-aware matrices which represent the edge information of local region.

\subsubsection{Feature extractor}
\begin{figure}[!htbp]
\centering
\vspace{-0.4cm}  
\includegraphics[width=0.7\textwidth]{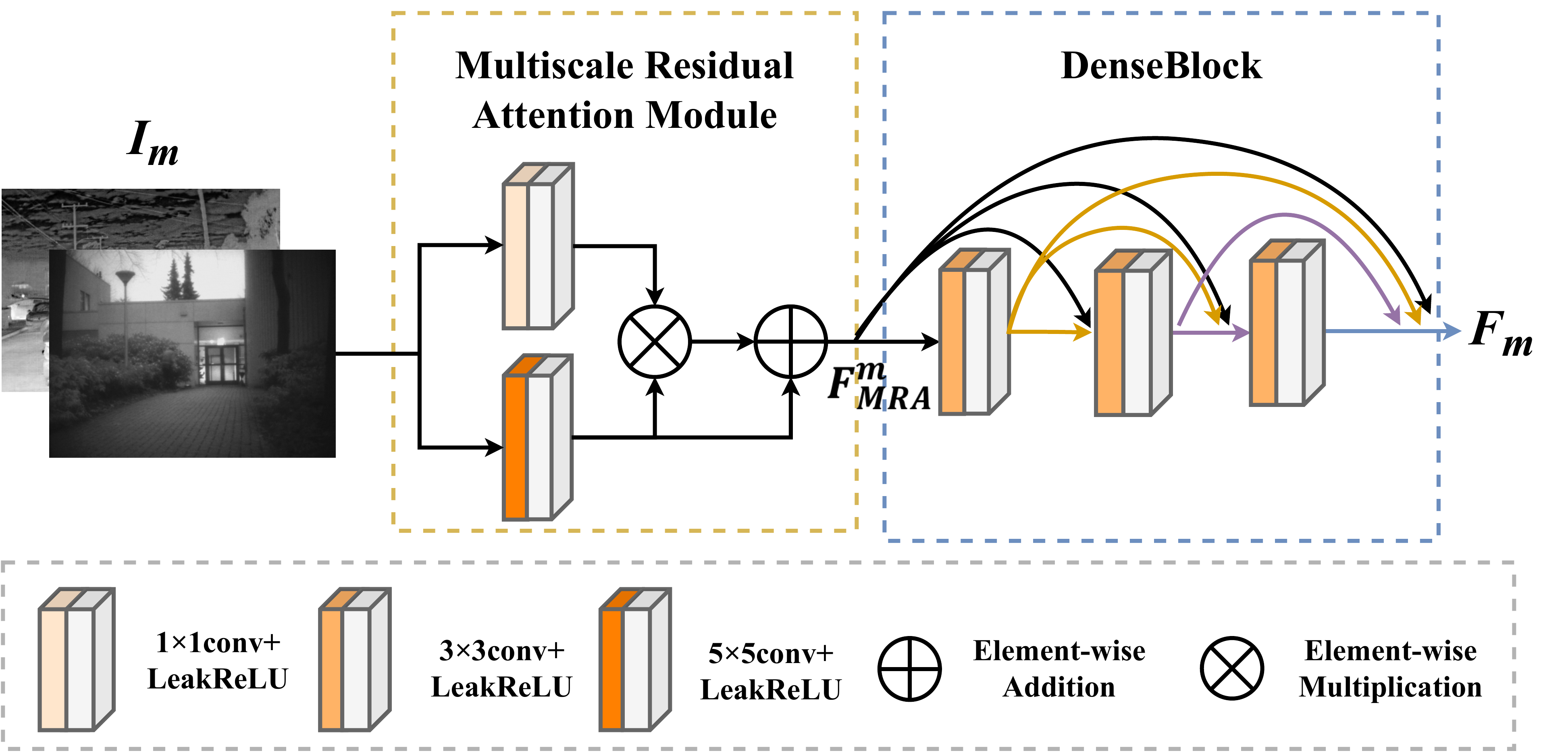}
\vspace{-0.3cm}
\caption{The detailed structure of feature extractor.} \label{fig4}
\vspace{-0.6cm} 
\end{figure}

In this section, we introduce the multi-scale residual attention network~\cite{fu2023mdranet} into the feature extractor to obtain richer image features (Fig.~\ref{fig4}).

In Fig.\ref{fig4}, the MRA contains several convolution layers which have different kernel size ($3\times 3$ and $5\times 5$), these layers are utilized to extract multi-scale features to enhance the image features. Then the DenseBlock is applied into feature extractor to preserve abundant detail information.
Specifically, we first extract the features of source images through the MRA module,
\vspace{-0.1cm}
\begin{equation}
F_{MRA}^{m}=conv_{5\times5}(I_{m})\oplus(conv_{5\times5}(I_{m})\otimes conv_{1\times1}(I_{m})), \quad m\in \left\{ir, vi \right\}
\end{equation}
where $conv_{n\times n}\left( \cdot \right)$ represents several convolutional operations with the kernel
size of $n\times n$, $\oplus$ refers to the element-wise summation, $\otimes$ indicates the element-wise multiplication. $F_{MRA}^{ir}$ and $F_{MRA}^{vi}$ are the features of the infrared and visible images obtained by the MRA module.

Then, we send these features into the DenseBlock, which is formulated as:
\begin{equation}
F_{m}=N_{Dense}(F_{MRA}^{m}), \quad m\in \left\{ir, vi \right\}
\end{equation}
where $N_{Dense}$ represents the DenseBlock.
$F_{ir}$ and $F_{vi}$ are the obtained rich features of source images.

\subsubsection{Fusion strategy}
In this part, the edge-aware matrices is utilized as the weights to enhance the multi-modal features, in which the attention and concatenation are applied.

Firstly, we combine the rich features extracted by the feature extractor together. It is specifically denoted as:
\begin{equation}
F_f=concat\left( (F_{ir}\oplus(F_{ir}\otimes E_{s})),(F_{vi}\oplus(F_{vi}\otimes E_{d})) \right),
\end{equation}
where $concat\left( \cdot \right)$ represents the concatenation in the channel dimensions.

\subsubsection{Image reconstructor}
\begin{figure}[!htbp]
\centering
\vspace{-0.3cm}
\includegraphics[width=0.8\textwidth]{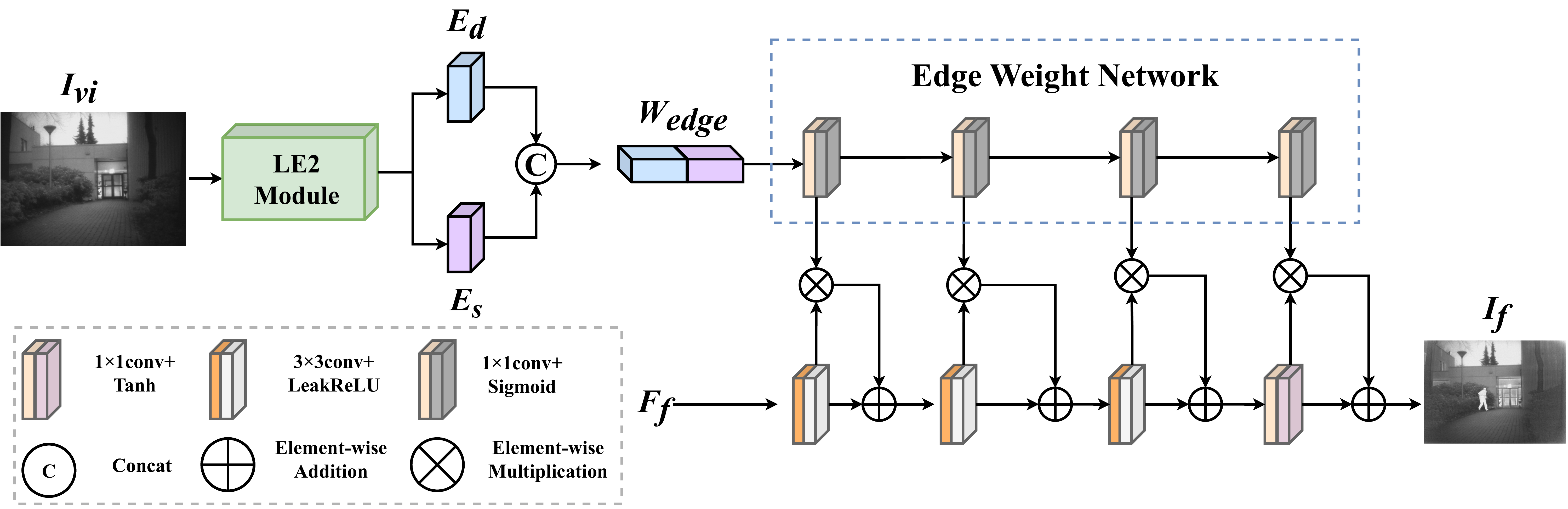}
\vspace{-0.2cm}
\caption{The framework of the image reconstructor.} \label{fig5}
\vspace{-0.6cm}
\end{figure}

As shown in Fig.~\ref{fig5}, in the image reconstruction process, we use a set of enhancement weights generated by the LE2 module to guide this process.

We use $F_{f}^{i}$ and $W_{edge}^{i}$ to denote the features reconstructed by the $i$th convolutional layer from the $F_f$ and $W_{edge}$, respectively.
The above image reconstruction process is defined as:

\begin{equation}
\begin{aligned}
F_{f}^{i}=F_{f}^{i-1}\oplus&(F_{f}^{i-1} \otimes conv_{1\times1}(W_{edge}^{i-1}))\\
s.t. \quad W_{edge} & =concat(E_d,E_b).
\end{aligned}
\end{equation}
\par
\subsection{Loss function}
A well-designed loss function enables different modules to learn specific information and make the network suitable for a specific fusion task.
To make our network extract more local edge features and detail information, our fusion loss is formulated as follows,
\begin{equation}
L_{fusion}=\lambda_1 \cdot L_{ssim}+\lambda_2 \cdot L_{region}+\lambda_3 \cdot L_{texture}.
\label{loss}
\end{equation}
where the $\lambda_1$, $\lambda_2$ and $\lambda_3$ indicate the balance parameters.

Specifically, the SSIM loss function is obtained by the following formulation:
\begin{equation}
L_{ssim}=1-SSIM\left( O,I \right)
\end{equation}
where $SSIM(\cdot)$ represents the structural similarity operation.

Unfortunately, for the local edge features, the structural similarity loss function is not promising.
To this end, we innovatively design a pixel intensity loss function based on the local regions, which is defined as follows,
\begin{equation}
L_{region}=\frac{1}{HW} \Vert I_f - max(I_{reg}^{ir}, I_{reg}^{vi}) \Vert_1,
\end{equation}
where $H$ and $W$ are the height and width of the source images.
$I_{reg}^{ir}$ and $I_{reg}^{vi}$ are image features extracted from the $3\times3$ regions with the average operation.

Besides, we expect the fused image to preserve salient texture details of source images. 
Thus, we specially introduce the texture loss, which is defined as follows,
\begin{equation}
L_{texture}=\frac{1}{HW} \Vert  \left| \nabla I_f \right| - max(\left| \nabla I_{ir} \right|, \left| \nabla I_{vi} \right|) \Vert_1,
\end{equation}
where $\nabla$ denotes the Sobel operator.

In summary, the design of our loss function is based on the idea of preserving meaningful information from source images.

\subsection{Network architecture} 
\subsubsection{Local edge enhancement network}
The specific structure of the local edge enhancement module is shown in Fig.~\ref{fig3}, which has six convolutional layers.
The kernel of the first convolutional layer is $5\times5$, and the kernels of the remaining five layers is $3\times3$.
Leaky Rectified Linear Unit (LReLU) activation function is used in all the convolutional layers.

\subsubsection{Feature extractor}
The feature extraction network consists of the multi-scale residual attention module and the DenseBlock.
The MRA module is composed of a $1\times1$ convolutional layer with the LReLU activation function and a $5\times5$ convolutional layer.
This module can fully extract the multi-scale image features.\par

The output of the MRA module will be the input of the DenseBlock, which contains three $3\times3$ convolutional layers, and the output of each layer is cascaded with the input of the next layer.
 Compared with the general networks, the introduction of the DesnBlock enables our network to obtain richer image features.
 \footnote{More details of network architecture, please refer to our supplementary material.}
 
\subsubsection{Image reconstructor}
In our approach, the image reconstructor is guided by a set of enhancement weights.
Four $1\times1$ convolutional layers are utilized to capture the edge information.
The image reconstructor contains three convolutional layers with the kernel size of $3\times3$ and one $1\times1$ layer used to decrease the channels. LReLU is used as the activation function for all convolutional layers in the image reconstructor, but the activation function of the last layer is the Tanh~\cite{ma2020ddcgan}.

\section{Experimental results and analysis}
In this section, we first introduce the details of our implementation configurations.
After that, we compare our method with several SOTA image fusion algorithms by performing qualitative and quantitative experiments.
In addition, we analyze the the edge features.
Finally, some ablation studies are conducted to verify the effectiveness of the proposed modules.
\subsection{Experimental settings}
To comprehensively evaluate the proposed algorithm, we perform qualitative and quantitative experiments on the MSRS dataset~\cite{tang2022piafusion} with all the 361 image pairs, the RoadScene dataset~\cite{xu2020u2fusion} with randomly selected 44 image pairs and the LLVIP dataset~\cite{jia2021llvip} with randomly selected 44 image pairs.
We compare our method with six state-of-the-art (SOTA) approaches, including DenseFuse~\cite{li2018densefuse}, FusionGAN~\cite{ma2019fusiongan}, U2Fusion~\cite{xu2020u2fusion}, PIAFusion~\cite{tang2022piafusion}, SwinFusion~\cite{ma2022swinfusion} and MUFusion~\cite{cheng2023mufusion}.
The implementations of these approaches are publicly available.\par
Five statistical evaluation metrics are selected in the quantitative experiments, including standard deviation (SD), entropy (EN), mutual information (MI), the sum of correlations of differences (SCD) and $Q_{abf}$.
SD reflects the visual effect of the fused image.
EN is used to represent the image detail retention.
MI measures the amount of information transferred from the source images to the fused image.
SCD reflects the level of correlation between the information transmitted to the fused image and corresponding source images.
$Q_{abf}$ measures the amount of edge information.
Moreover, a fusion algorithm with larger SD, EN, MI, SCD and $Q_{abf}$ indicates better fusion performance.

Our model is trained on the MSRS dataset, which contains 1,444 resized $256\times256$ image patches.
For the RGB images, we first convert the visible images to the YCbCr color space.
Then, the Y channel of the visible images and the infrared images are fused by our proposed method.
Finally, the fused image is converted back to the RGB color space via concatenating the Cb and Cr channels of the visible images.

The batch size and epochs are set as 30 and 4, respectively. The hyper-parameters $\lambda_1$, $\lambda_2$, and $\lambda_3$ in the Equation.~\ref{loss} are set as 3, 7, and 49, respectively. The model parameters are updated by using the Adam optimizer with the learning rate first initialized to 0.001 and then decayed exponentially. 
All the involved experiments are conducted on an NVIDIA RTX 3090Ti GPU and Intel Core i7-10700 CPU.

\subsection{Comparative experiments}
In order to comprehensively evaluate the performance of our method, we compare our method with other six SOTA methods on the MSRS and the LLVIP datasets\footnote{For more experiments, please refer to our supplementary material.}. 
\begin{figure}[!htbp]
\vspace{-0.3cm}
\centering
\includegraphics[width=0.95\textwidth]{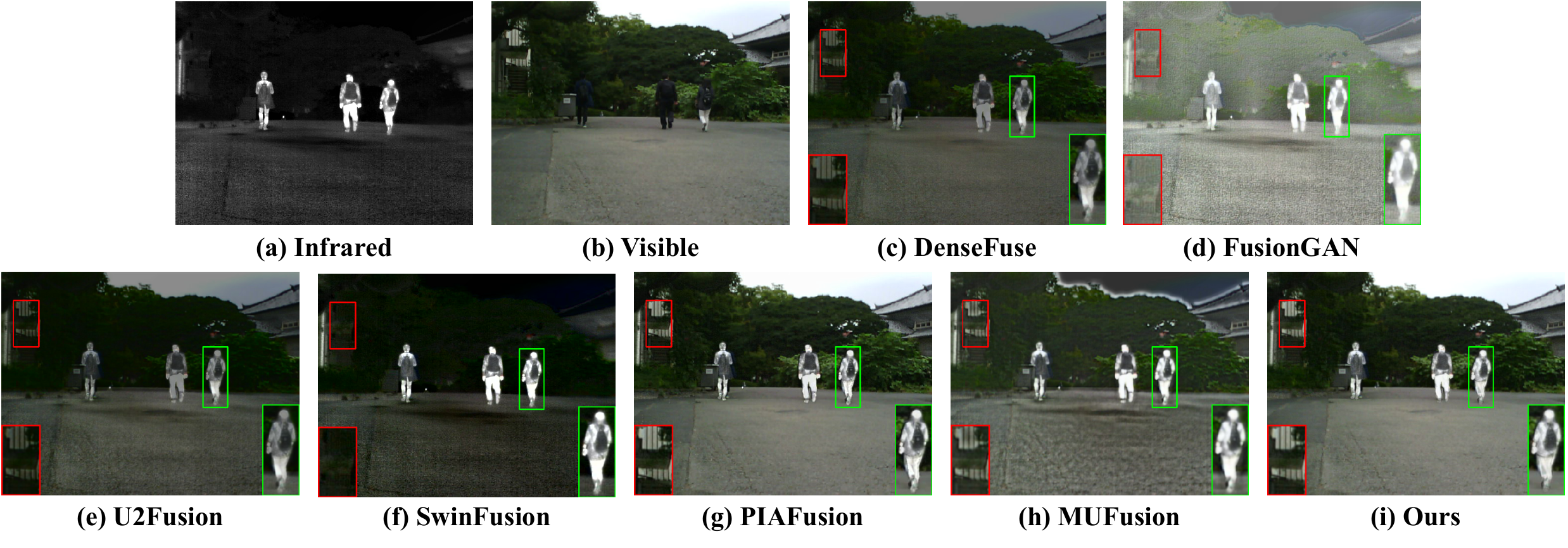}
\vspace{-0.4cm}
\caption{Qualitative comparison of our method with six SOTA methods on the MSRS dataset. 
} \label{fig6}
\vspace{-1.2cm}
\end{figure}

\subsubsection{Qualitative results}
 Qualitative experiments performed on the MSRS dataset are shown in Fig.~\ref{fig6}. As shown in the red highlighted regions, SwinFusion and FusionGAN are barely able to observe the details of the fence.
 Moreover, DenseFuse and U2Fusion weaken the target details. 
 For the MUFusion, it blurs the edges of the fence.
 Only our method and PIAFusion can better preserve the detail information.
 As illustrated in the green box, there is a mismatch issue between target and overall image light condition in FusionGAN, SwinFusion and MUFusion. 
 In addition, DenseFuse and U2Fusion weaken the infrared targets, resulting in suboptimal fusion results.\par
 \begin{figure}[!htbp]
\vspace{-0.3cm}
\centering
\includegraphics[width=0.95\textwidth]{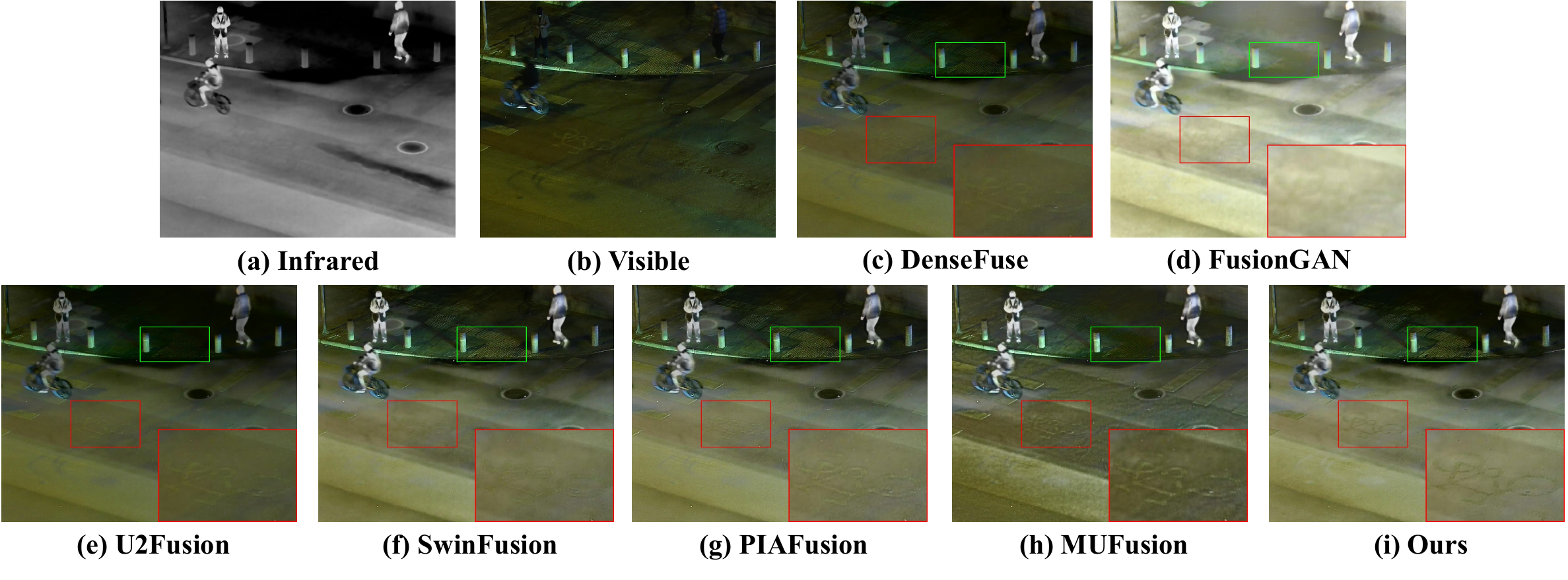}
\vspace{-0.4cm}
\caption{Qualitative comparison of our method with six SOTA methods on the LLVIP dataset. } \label{fig7}
\vspace{-0.6cm}
\end{figure}
The visualization results of different methods on the LLVIP dataset are shown in Fig.~\ref{fig7}.
The selected nighttime images can demonstrate the superiority of our approach.
As shown in the green box, DenseFuse, FusionGAN, U2Fusion, and MUFusion fail to clearly show the sidewalk in the dark, and FusionGAN fails to retain the sharp edges of the targets, suffering from the overall blurred issues.
Only our method, SwinFusion and PIAFusion preserve the texture information of the sidewalks well.
Besides, as shown in the red box, only our method can clearly present the bicycle mark.

\subsubsection{Quantitative results}
The quantitative results of five metrics on the MSRS dataset are presented in Table~\ref{tab2} (a).
The highest performance on SD proves that our fusion results are more in line with human visual perception.
Moreover, the best EN means that our fused image contains more meaningful information.
In addition, our method only follows PIAFusion by a narrow margin in SCD and $Q_{abf}$, which indicates that our fusion results can also contain more realistic information and edge information.
For the MI, although the advantage is not particularly significant, our method can effectively remove the artifacts in the fused images.
As a result, the amount of information transferred from the source image into the fused image is reduced.
Thus, it is reasonable that our performance on MI has lower ranking.\par

The comparison results of different methods on the LLVIP dataset are shown in Table~\ref{tab2} (b).
Our method ranks first on the metric of SD, which indicates that our method has satisfactory visual effects.
Although our method has the second best performance on EN and MI, the margin between our method and the best is tiny.
It demonstrates that our fusion results can contain more edge information and structure information.
In addition, our method only follows SwinFusion and PIAFusion by a narrow margin on SCD and $Q_{abf}$.
This is due to the reason that our method will sacrifice part of the information and reduce the noise when extracting the edge features of the local region, so as to produce a visually pleasing fused image. \par

\begin{table}[!htbp]
\centering
\vspace{-0.7cm}   
\caption{Quantitative results from MSRS and LLVIP datasets. (\textbf{Bold}: Best, \textcolor{red}{Red}: Second Best, \textcolor{blue}{Blue}: Third best)
}
\label{tab2}
\vspace{-0.7cm}
\end{table}
\begin{minipage}{\textwidth}\begin{minipage}[t]{0.48\textwidth}
\makeatletter\def\@captype{table}
\scalebox{0.7}{
\begin{tabular}{cccccc} 
\midrule 
\centering
Methods& SD  &  EN & MI & SCD & $Q_{abf}$ \\
\midrule 
DenseFuse & \textcolor{blue}{7.4237}   & 5.9340  & 2.5905  &  1.2489 &0.3572 \\
\midrule 
FusionGAN	& 7.1758     & \textcolor{blue}{5.9937}   &  1.4315  & 0.3129  & 0.2110 \\
\midrule 
U2Fusion	 & 6.8217    & 5.5515   &  2.3242  & \textcolor{blue}{1.2955}  &0.2972  \\
\midrule 
SwinFusion	& 6.0518     & 5.2846   & \textbf{4.1432}   & 0.5631  & 0.2942  \\
\midrule 
PIAFusion	 & \textcolor{red}{8.2822}     &\textcolor{red}{6.4971}    & \textcolor{red}{4.0022}   & \textbf{1.6421}  &\textbf{0.6486}  \\
\midrule 
MUFusion	 & 6.9233     & 5.9682   &1.6537    & 1.2548  &\textcolor{blue}{0.4110}  \\
\midrule 
Ours  & \textbf{8.3093}    & \textbf{6.5364}   &   \textcolor{blue}{3.7766} &\textcolor{red}{1.5803}  &\textcolor{red}{0.5996} \\
\bottomrule 
\end{tabular} 
}
\begin{center}
    (a) MSRS dataset
\end{center}
\end{minipage}
\begin{minipage}[t]{0.48\textwidth}
\makeatletter\def\@captype{table}
\scalebox{0.7}{
\begin{tabular}{cccccc} 
\midrule 
\centering
Methods& SD  &  EN & MI & SCD & $Q_{abf}$ \\
\midrule 
DenseFuse & 8.6065      & 6.4398   & 2.7610    & 1.0946   & 0.3426 \\
\midrule 
FusionGAN	&  \textcolor{red}{9.1167}      &  \textbf{6.9251}  & 1.9305    & 0.5557 & 0.2479 \\
\midrule 
U2Fusion	&8.5160        & 6.1217   &  3.0870   & 1.0862   &0.2983   \\
\midrule 
SwinFusion	& 8.7054       & \textcolor{blue}{6.7909}   & \textbf{3.8292}    & \textcolor{red}{1.3733}   &\textcolor{red}{0.5927}   \\
\midrule 
PIAFusion	 & \textcolor{blue}{8.9390}     & 6.7592   &  \textcolor{blue}{3.4146}   &  \textbf{1.4105}  & \textbf{0.6222}  \\
\midrule 
MUFusion	 & 8.3623       &  6.5243  & 2.5435    &1.0618    &0.4194   \\
\midrule 
Ours  &\textbf{9.2720}        & \textcolor{red}{6.8516}   &  \textcolor{red}{3.4509}   &  \textcolor{blue}{1.3242}  &\textcolor{blue}{0.4574}   \\
\bottomrule 
\end{tabular} 
}
\begin{center}
    (b) LLVIP dataset
\end{center}
\end{minipage}
\end{minipage}

\subsection{Ablation studies} 

\subsubsection{Edge features analysis}
\begin{figure}[!htbp]
\vspace{-0.5cm}
\centering
\includegraphics[width=0.9\linewidth]{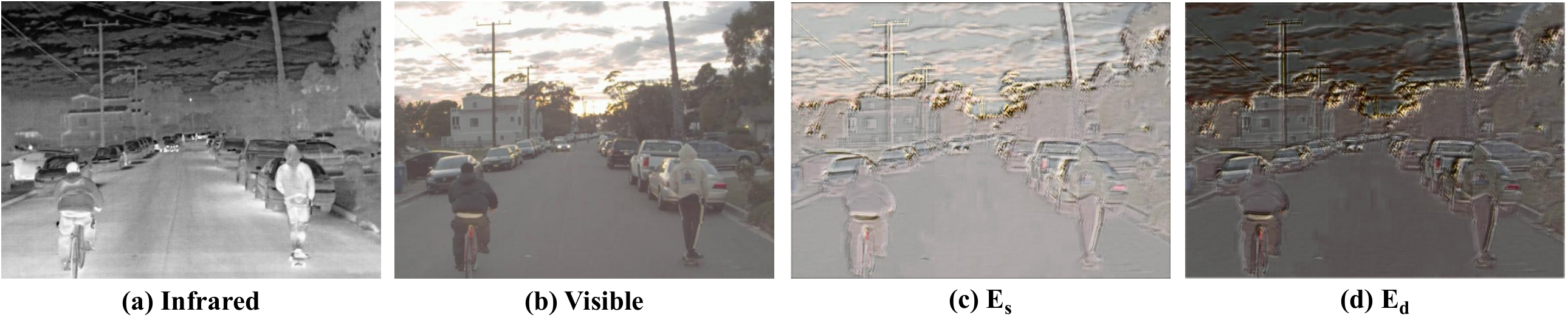}
\vspace{-0.4cm}
\caption{Visualized results of images and feature maps.} \label{fig8}
\vspace{-1.0cm}
\end{figure}

In Fig.~\ref{fig8}, we visualize a pair of feature maps generated by the LE2 module. 
We can clearly observe that the edge information are highlighted in the feature maps.
When dealing with the edge features of the local regions, we use a normalization formula, making two images complementary in brightness.
In our method, $E_d$ enhances the edge information and $E_s$ enhances other salient features.
As shown in the contrast of the trees and the sky in the edge feature map, the LE2 module extracts the meaningful information from the local edge regions.

\subsubsection{LE2 module analysis}
Considering the extraction of edge information in local regions, we design the LE2 module.
In ablation experiment, we design two removal methods, one is to remove the LE2 module during feature reconstruction, but retain it in the fusion layer (Fig.~\ref{fig9}(b) and Fig.~\ref{fig9}(g)).
The other approach is to remove the LE2 module completely (Fig.~\ref{fig9}(c) and Fig.~\ref{fig9}(h)).
From the green box, we can clearly find that they weaken the salient target.
Moreover, artifacts are also introduced in the enlarged red box in (g) and (h).
In addition, the results of the fusion layer with the LE2 is obviously better than the results without the LE2 module, which also demonstrates that the LE2 module plays an important role in guiding the image reconstruction. 

\subsubsection{MRA module analysis}
We introduce the MRA module into the DenseBlock to better preserve rich features from the source images.
To demonstrate its effectivenesss, we conduct ablation experiments on this module.
From (d), we can observer that the fused image without MRA module weakens the infrared salient target, and the sky also produces some artifacts.
From (i), more artifacts occurring on the door in the red box.
Moreover, the thermal target of the pedestrian is degraded in the green box.
In contrast, our MRA module can effectively reduce the artifacts and preserve abundant texture details.

\subsubsection{Analysis of the pixel intensity loss function}
\begin{figure}[!htbp]
\vspace{-0.4cm}
\centering
\includegraphics[width=\linewidth]{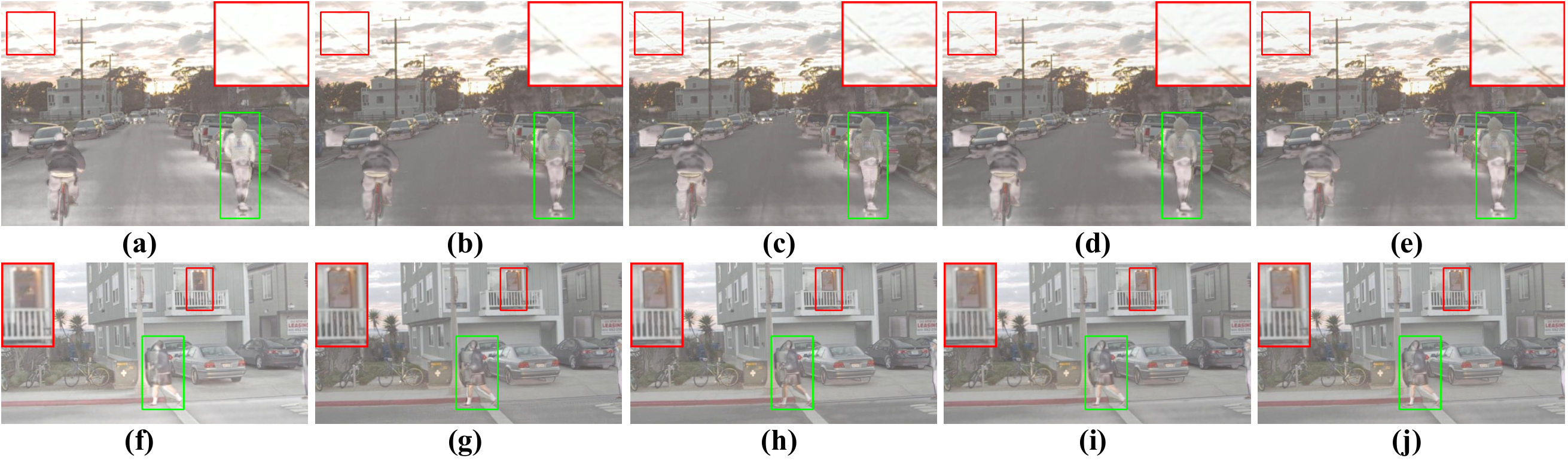}
\vspace{-0.7cm}
\caption{Visualized results of ablation studies. From left to right: fused results of our method, LE2 module in the fusion layer but not in the image reconstructor, without LE2 module completely, without MRA module and without pixel intensity loss.} \label{fig9}
\vspace{-1cm}
\end{figure}

Considering the different illumination conditions, we design the pixel intensity loss function based on local region to guide the network training, which makes the fused images contain more meaningful informaiton from the source images.
From Fig.~\ref{fig9}(e) and Fig.~\ref{fig9}(j), we can find that fused images weaken the salient target of the pedestrians.
Compared with other experimental settings, our loss function can make network focus on the local region edge information, which achieves better overall image quality.

\section{Conclusion}
In this work, we propose an infrared and visible image fusion network based on local edge enhancement module (LE2Fusion).
Specifically, we design a local edge enhancement module to enhance the edge information of the local regions and preserve salient information from the source images.
After that, it can generate a set of enhancement weights for guiding the feature fusion and image reconstruction processes.
To better guide the network training, we design a novel pixel intensity loss function based on the local regions, which enables our fusion results to maintain high correlation with source images in perspectives of the pixel intensities and the structure information.
Moreover, we introduce the multi-scale residual attention module into the DenseBlock to extract rich features.
Our method achieves comparable or better performance in terms of the visualization effects and quantitative evaluation.
Ablation experiments demonstrate the effectiveness of different components of the proposed method.

\subsubsection{Acknowledgements} This work was supported by the National Social Science Foundation of China(21$\&$ZD166), the National Natural Science Foundation of China (62202205), the Natural Science Foundation of Jiangsu Province, China(BK20221535), and the Fundamental Research Funds for the Central Universities (JUSRP123030).

%
%

\bibliographystyle{splncs04}
\bibliography{main}

\end{document}